# Machine-learning-based methods for output-only structural modal identification

Dawei Liu, Zhiyi Tang, Yuequan Bao[*], Hui Li

*Key Lab of Intelligent Disaster Mitigation of the Ministry of Industry and Information Technology, Harbin Institute of Technology, Harbin, 150090, China*

*School of Civil Engineering, Harbin Institute of Technology, Harbin China, 150090*

**Abstract:**

In this study, we propose a machine-learning-based approach to identify the modal parameters of the output-only data for structural health monitoring (SHM) that makes full use of the characteristic of independence of modal responses and the principle of machine learning. By taking advantage of the independence feature of each mode, we use the principle of unsupervised learning, making the training process of the deep neural network becomes the process of modal separation. A self-coding deep neural network is designed to identify the structural modal parameters from the vibration data of structures. The mixture signals, that is, the structural response data, are used as the input of the neural network. Then we use a complex loss function to restrict the training process of the neural network, making the output of the third layer the modal responses we want, and the weights of the last two layers are mode shapes. The deep neural network is essentially a nonlinear objective function optimization problem. A novel loss function is proposed to constrain the independent feature with consideration of uncorrelation and non-Gaussianity to restrict the designed neural network to obtain the structural modal parameters. A numerical example of a simple structure and an example of actual SHM data from a cable-stayed bridge are presented to illustrate the modal parameter identification ability of the proposed approach. The results show the approach's good capability in blindly extracting modal information from system responses.

## 1. Introduction

The identification of structural modal parameters (frequency, mode shape, and damping ratio) is a classical inverse problem of structural dynamics. Modal parameters represent the dynamic characteristics of a structure, which are only related to the physical parameters and mechanical models of the structure itself. They are of great significance in structural health monitoring (SHM) and are the basis of structural damage identification, model updating, and safety assessment. In the last decades, modal analysis has received much more attention in the identification of linear systems. It decouples a complex multiple-degree-of-freedom system into simple modal superposition, which enables a description of the structure's vibration to be done more efficiently. Modal parameters, that is, modal frequency damping ratios and mode shapes, comprise the fundamental information of the dynamic characteristics of a structure. They describe how structures behave under external forces. Therefore, they contain vital information about a structure, can be used for structural damage detection, model updating, and structural safety assessment.

Traditional modal identification methods utilize the relationship between the input and output of the structures. This needs an ideal test situation in which excitation of the structure can be controlled or measured.[1] However, in many actual infrastructures, it is impossible to achieve the excitation, so output-only modal identification methods more significant advantages.[2] In the past several decades, many of these types of methods have emerged, involving the time domain, frequency domain, and time-frequency domain.[3] The Ibrahim time-domain (ITD) method[4] uses the displacement, velocity, or acceleration time-domain signal of the structural free vibration response to identify the modal parameters. The peak-picking (PP) method is based on the principle that the frequency response function appears as a peak near the natural frequency, and the power spectrum of the random response signal is used instead of the frequency response function. The time-series analysis method based on the autoregressive moving average (ARMA) model is a method of processing ordered randomly sampled data using a parametric model to obtain modal parameters. The stochastic subspace identification (SSI)[5] method is based on the identification method of discrete state-space equations in linear systems, which is suitable for stationary excitation. The natural excitation technique (NExT)[6] replaces the free vibration response or impulse response function in traditional time-domain modal analysis with a cross-correlation function between responses. The eigensystem realization algorithm (ERA)[7] uses the measured impulse response data or free-response data to form a Hankel matrix and uses singular value decomposition to find a minimum implementation of the system. Frequency-domain decomposition (FDD)[8,9] is an extension of the peak-picking method. The main idea is to perform singular value decomposition on the power spectrum of the response and decompose the power spectrum into a set of single-degree-of-freedom system power spectra corresponding to multiple modalities. Empirical mode decomposition (EMD)[10] is a signal decomposition method based on local features of signals. In recent years, blind source separation (BSS) has been successfully used in structural dynamics for, for example, modal identification.[11] Most of the BSS methods exploit four types of mathematical properties, namely mutual independence between the sources (also called independent component analysis (ICA)),[2,12,13] sparse representation of the sources (such as sparse component analysis (SCA)),[14–18] temporal structure of the sources[19–22] (such as second-order blind identification (SOBI)[20,21]), and the algorithm for multiple unknown signals extraction (AMUSE),[23] and the non-stationary of the sources.[24]

The modal identification methods, such as FDD, ERA, SSI, ICA, SCA, and SOBI, have been successfully applied in output-only modal identification. Brincker *et al*. used FDD to identify close modes with high accuracy, even in the case of strong noise contamination of the signals.[8] Caicedo *et al*. discussed the steps and parameters to perform an effective modal identification using the NExT and ERA.[25] Peeters *et al*. presented the use of SSI for in-operation modal analysis and changed the row space of the future outputs to be projected into the row space of the past reference outputs.[26] Gao *et al*. proposed an automated operational modal analysis approach based on SSI and clustering.[27] Kerschen *et al*. discussed the relation between the vibration modes of mechanical systems and the modes computed through a blind source separation technique, also known as ICA.[13] Yang *et al*. proposed an ICA-based time-frequency BSS framework that can perform modal identification of both lightly and highly damped structures.[2] Yang *et al*. also introduced SCA into modal identification, which proved to be simple and efficient in conducting reliable output-only modal identification even with limited sensors.[14] Most of the methods use the relationship between monitoring data and the structural system matrix to identify the modal parameters. However, the model order problem remains a challenge, which requires expert involvement and time-consuming computational overhead, and may be affected by measurement noise and non-stationary excitations.[1]

Structural modal parameter identification is a classical inverse problem of structural dynamics, and its essence is an optimization problem. The ability of an optimization algorithm determines the identification capacity for modal parameters. In recent years, machine learning has increasingly become an emerging and more effective method for various disciplines. Machine learning is a powerful fitting learning algorithm that can find a global minimum even if the objective function is non-convex and non-smooth.[28] Han *et al*. used deep learning to solve high-dimensional partial-differential equations, and the result showed that the method has better accuracy and cost-effectiveness under high-dimensional conditions.[29] Li *et al*. used a deep-learning method to solve partial-differential equation control problems,[30] which formulates the deep residual network (ResNet) as a control problem of a transport equation. Chen *et al*. introduced a new family of deep-neural-network (DNN) models in which the output of the network is computed by a black-box differential equation solver.[31] Sun *et al*. used a network of alternating direction method of multipliers (ADMM-Net) for fast magnetic resonance imaging (MRI),[32] and the result showed that it achieved high reconstruction accuracies with fast computational speed. Carleo *et al*. used an artificial neural network (ANN) to solve quantum mechanics problems and achieved high accuracy in describing a prototypical interaction spin model in one and two dimensions.[33]

Machine learning plays an increasingly important role in solving and optimizing mathematical models of civil engineering. Wei *et al*. used a deep reinforcement learning method to solve nonlinear differential equations[34] that can give solutions with high accuracy, and the process promises to become faster. Bao *et al*. proposed a neural network method to solve the data compressive sensing problem.[35] Research using machine learning to solve equations or optimization problems in civil engineering has just started, and other application studies, such as computer vision-based structural inspection,[36–40] data anomaly detection,[41,42] system prediction,[43] and lost-data recovery,[44] are becoming relatively common.

Inspired by machine learning to solve optimization problems in civil engineering, we propose using machine learning to solve modal equations to identify modal parameters. Some researchers have studied structural modal identification based on machine learning. Facchini *et al*.[45] used four-frequency-dependent indicators as the input of the neural network for output-only modal identification. The trained ANN in their work can identify the modal parameters of the structure, but their method cannot directly obtain the modal parameters from the structural response; it must calculate four indicators from the signal first. Thus, it does not realize the automatic identification of modal parameters using neural networks.

In this study, a machine-learning-based method for output-only structural modal identification is proposed, which makes full use of the characteristic of the independence of modal response. Upon integrating the independence features into the neural network, the accuracy and stability of structural modal identification are well improved.

## 2. Machine-learning method for structural modal identification

### 2.1. Principle of the method

According to structural dynamics theory, for an *n*-degree-of-freedom (DOF) linear time-invariant structure system, the governing equation of motion can be written as

$$\mathbf{M}\ddot{\mathbf{x}}(t) + \mathbf{C}\dot{\mathbf{x}}(t) + \mathbf{K}\mathbf{x}(t) = \mathbf{f}(t) \tag{1}$$

where $\mathbf{M} \in \Re^{n \times n}$ is the mass matrix, $\mathbf{C} \in \Re^{n \times n}$ the damping matrix, and $\mathbf{K} \in \Re^{n \times n}$ the stiffness matrix, $\mathbf{f}(t)$ the external force, $\mathbf{x}(t)$ the system response, and the dot denotes derivatives with respect to time.

The system response $\mathbf{x}(t) = [x_1(t), \ldots, x_n(t)]^T$ can be expressed using the modal expansion:

$$\mathbf{x}(t) = \mathbf{\Phi}\mathbf{q}(t) = \sum_{i=1}^{n} \varphi_i q_i(t) \tag{2}$$

where $\mathbf{\Phi} \in \Re^{n \times n}$ is the modal matrix that consists of the mode shape vector $\varphi_i \in \Re^n$, and $\mathbf{q}(t) = [q_1(t), \ldots, q_n(t)]^T$ is the modal response, i.e., modal coordinates.

In free vibration, that is $\mathbf{f}(t) = 0$, the $q_i(t)$ can be expressed as monotone exponentially decaying sinusoids:[13]

$$q_i(t) = u_i \exp^{-\zeta_i \omega_i t} \cos(\omega_{di} t + \theta_i), i = 1, \ldots, n \tag{3}$$

where $\omega_i$ and $\zeta_i$ are natural frequencies and damping ratios; $\omega_{di}$ is the damped natural frequency, $\omega_{di} = \omega_i \sqrt{1 - \zeta_i^2}$; $u_i$ and $\theta_i$ are constants under corresponding initial conditions.

In random vibration, $q_i(t)$ can be approximated as modulated by a random envelope function $e_i(t)$,[1,13]

$$q_i(t) \cong e_i(t) u_i \exp^{-\zeta_i \omega_i t} \cos(\omega_{di} t + \theta_i), \quad i = 1, \ldots, n \tag{4}$$

Therefore, the problem is how to obtain the modal responses $\mathbf{q}(t)$ and mode shapes $\mathbf{\Phi}$ from the system response $\mathbf{x}(t)$. Inspired by the BSS and DNN, we formalize the modal identification problem of Eq. (2) into a DNN and use the powerful optimization ability of the DNN to solve Eq. (2) to separate the modal response $\mathbf{q}(t)$ and mode shapes $\mathbf{\Phi}$ from the system response. In designing the neural network, the independent characteristic of modal responses is used as the constraint for the loss function. Figure 1 illustrates the framework of the proposed method. Different from a regular black-box neural network, the designed neural network is interpretable for each layer. As shown in the figure, the raw time response data of the structure are pre-processed with filtering and denoising. Then, the data are fed into the neural network to obtain the modal responses and mode shapes. After obtaining the mode responses, the frequencies and damping ratios are obtained by traditional power spectral density (PSD) and curve-fitting methods, respectively. The mode shapes are the weights of the last two layers.

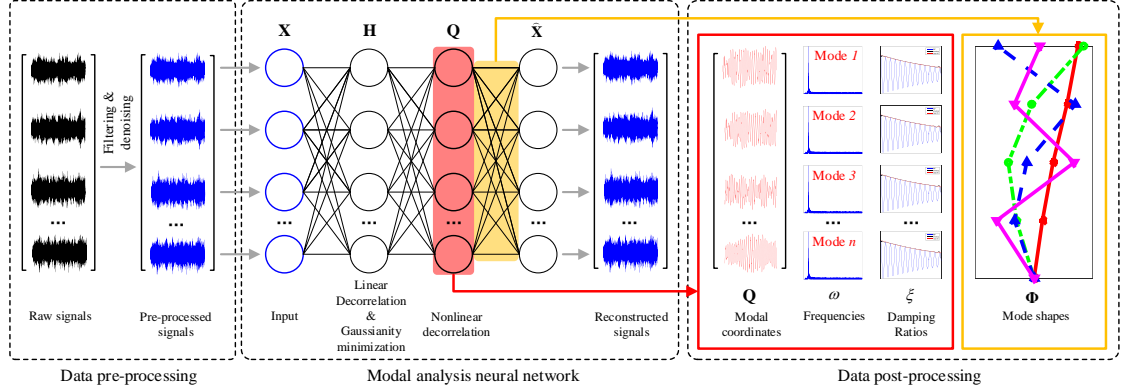

Figure 1 Flowchart of the method

## 2.2. Modal identification in DNN framework

### 2.2.1. Principle of independence

The main idea of the proposed method is the use of the designed DNN to separate the modal responses from the vibration response of structures. The method of using independent modal characteristics to identify modal parameters of a weakly damped system is valid when the natural frequencies are incommensurable (i.e., the natural frequencies cannot be in a simple integer ratio).[13] In the case of free vibration or random vibration of weakly damped system, the mixing matrix and the modal responses of the structure are one to one mapping. Therefore, the modal response can be seen as independent virtual sources and using the blind source separation methods to separate it[2, 3, 13]. Under environmental excitation, the modal responses of a weakly damped system are statistically independent.[13] Usually, there are two aspects of measuring the independence, namely non-Gaussianity and uncorrelation.

*(1) Non-Gaussianity*

The idea of maximum non-Gaussianity comes from the central limit theorem, which means if the observation signal is a linear combination of several independent sources, then the observation signal is closer to the Gaussian distribution than the source signals; that is, the source signal's non-Gaussianity is stronger than the observation signal.[46] Therefore, optimal separation results can be obtained by maximizing the non-Gaussianity of the separation results.

There are two indicators that can measure non-Gaussianity, kurtosis, and negative entropy. However, kurtosis is sensitive to singular values, which means it is not a robust non-Gaussianity measurement index.

Hyvärinen proposed a more robust and faster way to approximate negative entropy,[46]

$$J(s_i) \approx \left[ E\{G(s_i)\} - E\{G(v)\} \right]^2 \qquad (5)$$

where $E\{\cdot\}$ is expectation operator, $s_i$ the output variable with a zero mean and unit variance, $v$ a Gaussian random variable with zero mean and unit variance variable, and $G$ a non-squared nonlinear function. In practice, we often choose the following functions:[12]

$$G_1(s) = \frac{1}{a_1} \log_2 \cosh(a_1 s) \tag{6}$$

$$G_2(s) = -\exp\left(-\frac{s^2}{2}\right) \tag{7}$$

$$G_3(s) = \frac{s^4}{4} \tag{8}$$

where $1 \leq a_1 \leq 2$, typically takes 1. $G_1$ is suitable for the coexistence of sub-Gaussian signal and super-Gaussian signal, $G_2$ is suitable for separating super-Gaussian mixed signals, $G_3$ is suitable for the separation of sub-Gaussian mixed signals, where vibration signals are sub-Gaussian mixed signals. Since the vibration signal is a sub-Gaussian signal, we use the $G_3$ function in this study. For the sub-Gaussian signal, when the $G_3$ function takes the minimum value, the Gaussianity property is the smallest, and the non-Gaussianity property is the largest.

In the method of separating signals by negative entropy, whitening is usually used as a means of data pre-processing, which makes it easier to obtain mutually independent components from the signal. The essence of whitening is to achieve a linear uncorrelated signal; that is, by a linear transformation $\mathbf{u}$, the variable becomes a linearly uncorrelated variable,[12] which is

$$E\{\mathbf{y}\mathbf{y}^T\} = \mathbf{I}, \ \mathbf{y} = \mathbf{u}\mathbf{s} \tag{9}$$

When the mean of the signal $\mathbf{s}$ is zero, the above formula becomes

$$E\{\mathbf{u}\mathbf{u}^T\} = \mathbf{I} \tag{10}$$

*(2) Uncorrelation*

Uncorrelation includes the linear and nonlinear uncorrelation of two variables, which represents the two independent variables

$$E\{s_1 s_2\} = E\{s_1\} E\{s_2\} \tag{11}$$

$$E\{f(s_1) g(s_2)\} = E\{f(s_1)\} E\{g(s_2)\} \tag{12}$$

where $s_1$ and $s_2$ are two variables, $f(\cdot)$ and $g(\cdot)$ are nonlinear transformation.

According to Eq. (2), the modal responses $q_i(t)$, $i = 1, \ldots, n$ comprise uncorrelation and have non-Gaussianity. Therefore, these characteristics can be used to separate the modal responses from the original vibration signals of structures. To increase the modal response separation effectively, both the uncorrelation and non-Gaussianity are considered in the proposed method; however, the traditional BSS methods[12,13] are only used to consider the non-Gaussianity.

*2.2.2. Design of DNNs*

The uncorrelation and non-Gaussianity, that is, negative entropy, of the vibration signals are employed and embedded into the neural network as the constraint functions. The architecture of the designed network for modal identification is shown in Figure 2. There are four steps: (1) feeding the time response data of the structure into the network and randomly initializing the weights; (2) performing forward propagation and calculating the loss function of each layer to get the total loss function; (3) performing backward propagation and using optimization algorithms to update network weights; and (4) iteratively updating the weights and debugging hyperparameters until the network converges.

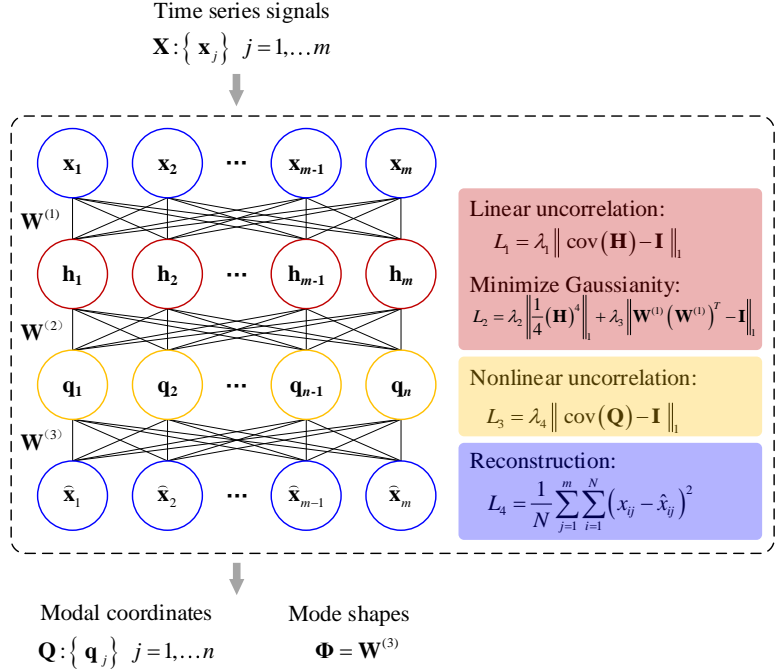

Figure 2 Architecture of the network for modal identification

The entire loss function of the designed neural network is

$$
\begin{aligned}
L &= L_1 + L_2 + L_3 + L_4 \\
&= \lambda_1 \| \text{cov}(\mathbf{H}) - \mathbf{I} \|_1 + \lambda_2 \left\| \frac{1}{4}(\mathbf{H})^4 \right\|_1 + \lambda_3 \| \mathbf{W}^{(1)} (\mathbf{W}^{(1)})^T - \mathbf{I} \|_1 \\
&\quad + \lambda_4 \| \text{cov}(\mathbf{Q}) - \mathbf{I} \|_1 + \frac{1}{N} \sum_{j=1}^{m} \sum_{i=1}^{N} (x_{ij} - \hat{x}_{ij})^2 \\
&= \lambda_1 \max_{1 \le k \le m} \sum_{j=1}^{m} |h_{jk} - \mathbf{I}_{jk}| + \lambda_2 \max_{1 \le j \le m} \sum_{i=1}^{N} \left| \frac{1}{4}(h_{ij})^4 \right| + \lambda_3 \max_{1 \le w \le m} \sum_{u=1}^{m} \left| \sum_{v=1}^{m} (w_{uv})^2 - \mathbf{I}_{uw} \right| \\
&\quad \lambda_4 \max_{1 \le s \le n} \sum_{r=1}^{n} |q_{rs} - \mathbf{I}_{rs}| + \frac{1}{N} \sum_{j=1}^{m} \sum_{i=1}^{N} (x_{ij} - \hat{x}_{ij})^2
\end{aligned} \tag{13}
$$

The input to the neural network is the structural vibration response data $\mathbf{X}$ measured by sensors after pre-processing, for example, filtering and denoising, and the number of input neurons is equal to the number of sensor channels, which is

$$
\mathbf{X} = \begin{pmatrix} x_{11} & \cdots & x_{1m} \\ \vdots & x_{ij} & \vdots \\ x_{N1} & \cdots & x_{Nm} \end{pmatrix}, \mathbf{X} \in \Re^{N \times m} \tag{14}
$$

where $N$ are the sample points of time-domain data and $m$ is the number of sensor channels.

Layer 2 of the neural network is restricted by linear uncorrelation and minimized Gaussianity. Therefore, the result of layer 2's covariance matrix is restricted close to the identity matrix, which ensures uncorrelated linear constraints. The constraint that minimizes the Gaussianity of the signal has two parts, including whitening [third term in Eq. (13)] and non-Gaussianity functions [second term in Eq. (13)] to obtain minimum values:

$$L_1+L_2 = \lambda_1 \| \text{cov}(\mathbf{H}) - \mathbf{I} \|_1 + \lambda_2 \left\| \frac{1}{4}(\mathbf{H})^4 \right\|_1 + \lambda_3 \| \mathbf{W}^{(1)} \left(\mathbf{W}^{(1)}\right)^T - \mathbf{I} \|_1$$
$$= \lambda_1 \max_{1 \leq k \leq m} \sum_{j=1}^{m} |h_{jk} - \mathbf{I}_{jk}| + \lambda_2 \max_{1 \leq j \leq m} \sum_{i=1}^{N} \left|\frac{1}{4}(h_{ij})^4\right| + \lambda_3 \max_{1 \leq w \leq m} \sum_{u=1}^{m} \left|\sum_{v=1}^{m}(w_{uv})^2 - \mathbf{I}_{uw}\right| \quad (15)$$

where $\mathbf{H}$ is the result of layer 2, $\text{cov}(\cdot)$ the covariance between different variables, $\mathbf{I}$ the identity matrix and $\|\cdot\|_1$ the $l_1$ norm, $\mathbf{W}^{(1)}$ the weights between layer 1 and layer 2; $\lambda_1$ $\lambda_2$, and $\lambda_3$ are a constant between 0 and 1.

Layer 3 of the neural network is restricted by nonlinear uncorrelation. Layer 3 has an activation function (tanh function is used in this neural network) to apply a nonlinear transformation. Then the result of layer 3's covariance matrix is restricted close to the identity matrix, which ensures nonlinear uncorrelated constraints:

$$L_3 = \lambda_4 \| \text{cov}(\mathbf{Q}) - \mathbf{I} \|_1 = \lambda_4 \max_{1 \leq s \leq n} \sum_{r=1}^{n} |q_{rs} - \mathbf{I}_{rs}| \quad (16)$$

where the modal response matrix $\mathbf{Q}$ is the result of layer 3, $\lambda_4$ is a constant between 0 and 1, $n$ is the number of modal responses.

Layer 4 of the neural network is used for reconstruction of the original input data, which ensures the results of the layer 3 are the modal responses, and the weights of layers 3 and 4 are the mode shapes. A loss function is used to restrict layer 4 to reconstructing the input:

$$L_4 = \frac{1}{N} \sum_{j=1}^{m} \sum_{i=1}^{N} (x_{ij} - \hat{x}_{ij})^2 \quad (17)$$

where $x_{ij}$ is the input data and $\hat{x}_{ij}$ the reconstruction data.

For the optimization of the loss function, the RMSProp[47] algorithm is employed because of its good optimization ability under non-convex conditions. The neural network weight update method is as follows:

$$\begin{aligned} g_{t+1} &= \nabla_w L(w_t) \\ G_{t+1} &= \lambda G_t + (1-\lambda) g_{t+1} \otimes g_{t+1} \\ w_{t+1} &= w_t - \frac{\tilde{\gamma}}{\sqrt{G_{t+1} + \varepsilon}} \otimes g_{t+1} \end{aligned} \quad (18)$$

where $t$ is the step of iteration, $L(w_t)$ the loss function at the step of iteration $t$,

$\nabla_w L(w_t)$ the gradient of $L$ with respect to $w$, $g$ the calculating gradient, $G$ the cumulative square gradient, $\lambda$ the decay rate, $\otimes$ the elementwise square, $w$ the network weight, $\tilde{\gamma}$ the global learning rate, and $\varepsilon$ a small constant.

After the neural network training is completed, the new structural response data can be input into the neural network for estimating the modal parameters. According to the principle of independence of BSS, the obtained most independent components from structural responses data are the most closet to the modal responses. The output of layer 3 meet the maximum independence; therefore, Layer 3's results are the optimal estimation of the modal responses:

$$\hat{\mathbf{Q}} = \begin{pmatrix} q_{11} & \cdots & q_{1n} \\ \vdots & q_{ij} & \vdots \\ q_{N1} & \cdots & q_{Nn} \end{pmatrix}, \quad \hat{\mathbf{Q}} \in \Re^{N \times n} \tag{19}$$

where each column of $\hat{\mathbf{Q}}$ is an estimation of modal response.

According to the Eq. (2), the product of modal responses and mode shapes coefficients are the system responses (i.e., the input of the fourth layer in the designed network); therefore, the weights between layer 3 and 4 are the optimal estimation of mode shapes:

$$\hat{\mathbf{\Phi}} = \mathbf{W}^{(3)} = \begin{pmatrix} w_{11}^{(3)} & \cdots & w_{1m}^{(3)} \\ \vdots & w_{uv}^{(3)} & \vdots \\ w_{n1}^{(3)} & \cdots & w_{nm}^{(3)} \end{pmatrix}, \quad \mathbf{W}^{(3)} \in \Re^{n \times m} \tag{20}$$

where each row is the estimation of mode shape.

*2.2.3. Modal parameters*

As shown in Eq. (20), the mode shapes are the weights of layers 3 and 4 of the neural network and can be directly obtained from the neural network. After obtaining the optimal estimation of the modal responses $\hat{\mathbf{Q}}$, the modal frequencies and damping ratios can be determined using traditional modal analysis methods. By picking the peak of the PSD of modal responses, the modal frequencies are determined. Using the free vibration attenuation method, the damping ratio can be determined, where the free vibration curves are obtained from the modal responses by the random decrement technique (RDT).

## 3. Numerical simulations

To validate the proposed machine-learning-based modal identification method, numerical simulations are conducted on a 4DOF linear time-invariant spring-mass damped model (Figure 3).

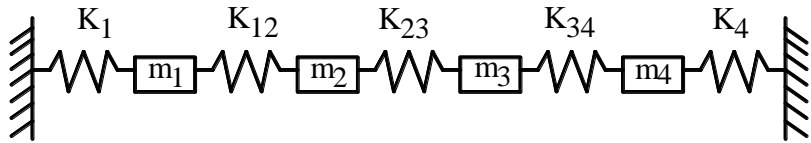

Figure 3 Four DOFs linear spring-mass damped system

The structural matrices are given as follows:

$$\mathbf{M} = \begin{bmatrix} 1 & & & \\ & 2 & & \\ & & 3 & \\ & & & 4 \end{bmatrix}, \quad \mathbf{K} = \begin{bmatrix} 1000 & -800 & 0 & 0 \\ -800 & 2400 & -1600 & 0 \\ 0 & -1600 & 4800 & -3200 \\ 0 & 0 & -3200 & 8000 \end{bmatrix}, \quad \mathbf{C} = \alpha \mathbf{M} + \beta \mathbf{K}$$

The method that uses independence is not suitable for use in high-damping cases because the modal responses do not satisfy the assumption of independence when the damping level is high. For undamped and lightly damped systems, the modal responses are approximately monotone sinusoids, and the system responses are linear mixtures of sinusoids, so the requirement of independence is met.[2]

In this numerical example, the mass proportional damping is set as the damping factors $\alpha = 0.1$, $\beta = 0$. The stationary white Gaussian noise is used to excite the system at the four DOFs and the Newmark-beta algorithm with a sampling frequency of 100 Hz to calculate the system responses. The system responses are listed in Figure 4, which shows the response data in the time domain, frequency domain, and time-frequency domain separately.

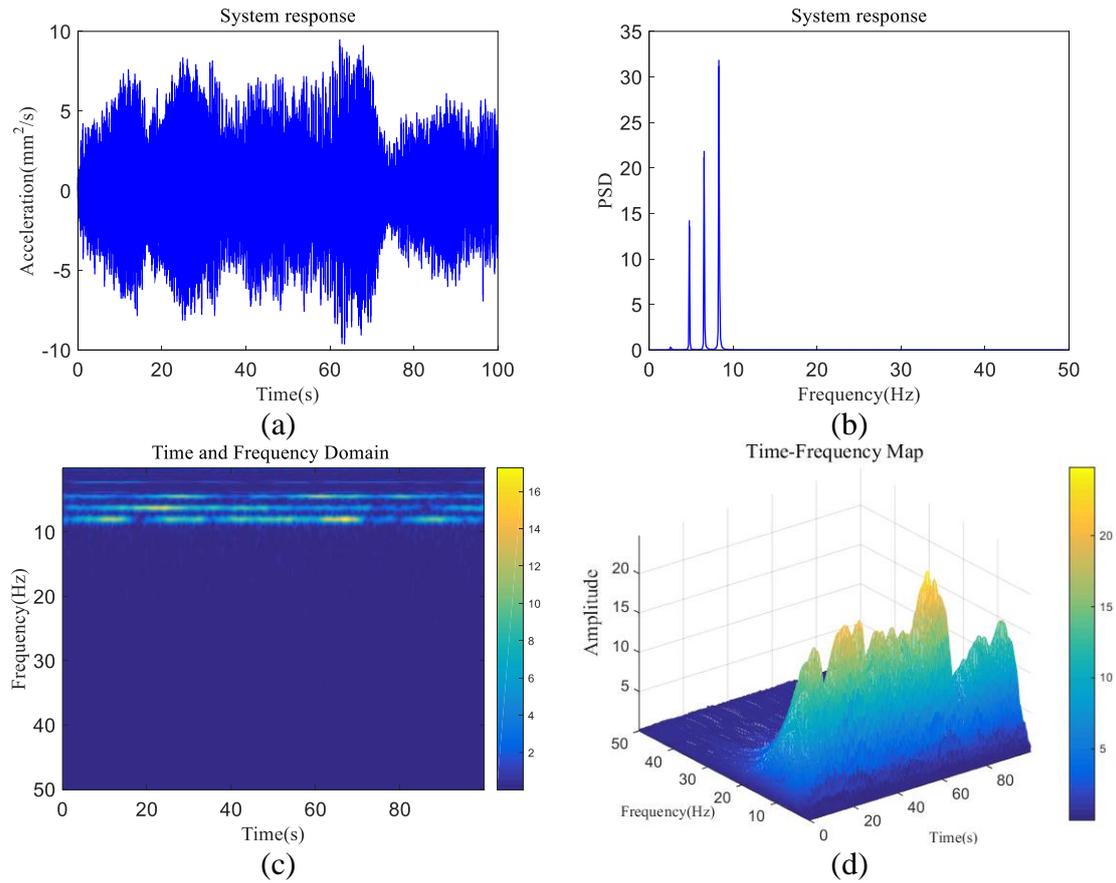

Figure 4 System response: (a) time response data, (b) PSD, (c) time-frequency analysis results, and (d) three-dimensional time-frequency map

The architecture of the designed neural network is shown in Figure 5. In the training stage, the system responses $\mathbf{x}_i \, (i=1,\ldots,4)$ with 1,000 data points are used as the input of the designed network. At the same time, all network weights are randomly initialized. Then, the loss function of each layer is calculated by the forward propagation algorithm to obtain the total loss function of Eq. (13); and the backpropagation is carried out to update the weights of each layer. Finally, weights updating and hyperparameters debugging are iteratively performed until the network converges. After the network converges, the results of the third layer $\hat{\mathbf{Q}}$ (i.e., the optimal estimation of the modal responses) and the weight values $\mathbf{W}^{(3)}$ between the third and fourth layers (i.e., the optimal estimation of mode shapes) can be extracted.

The learning rate, batch size, and training epochs of the neural network were 0.01,128 and 10,000, respectively. The training and convergence of the network are shown in Figure 5, which indicates that the convergence measured by the proposed loss function [Eq. (13)] becomes optimal at epoch 1,000.

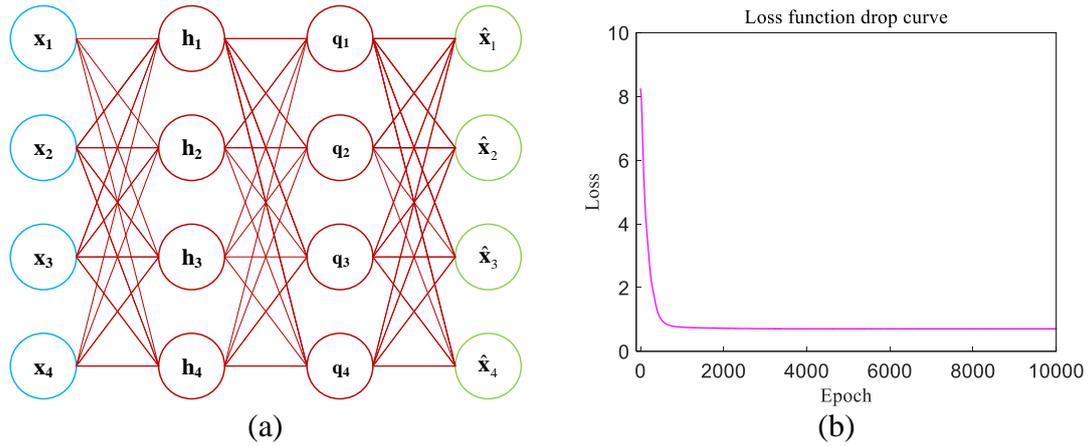

Figure 5 DNN architecture and training results: (a) architecture of DNN and (b) convergence in the training process

The normalized four modal responses estimated by the neural network are shown in Figure 6. The results in the time domain and frequency domain show the modal responses are accurately estimated.

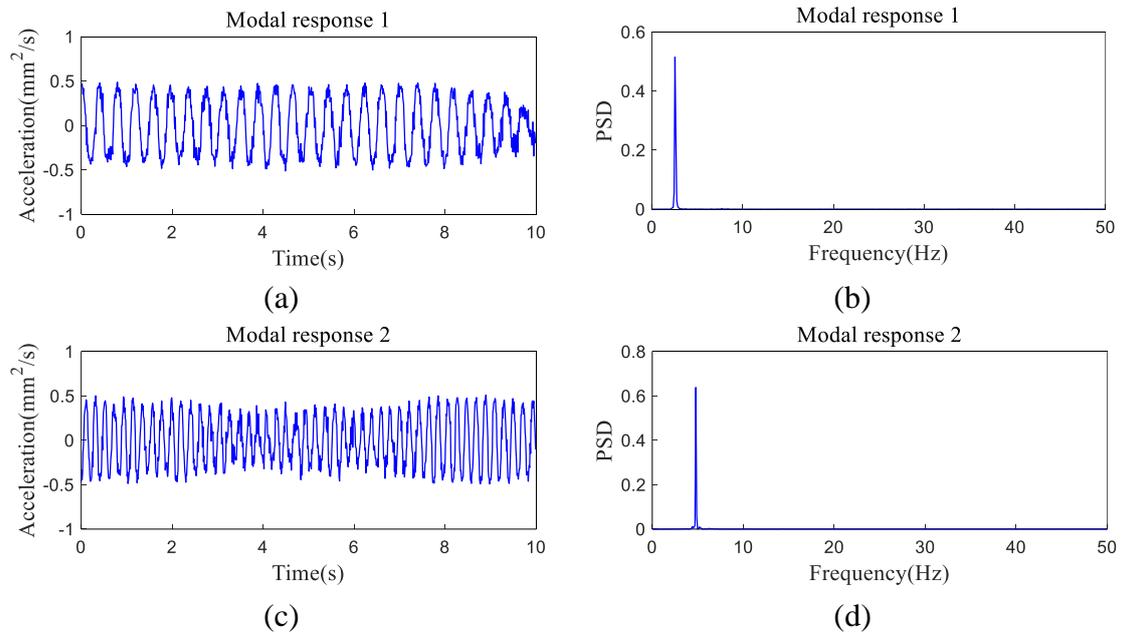

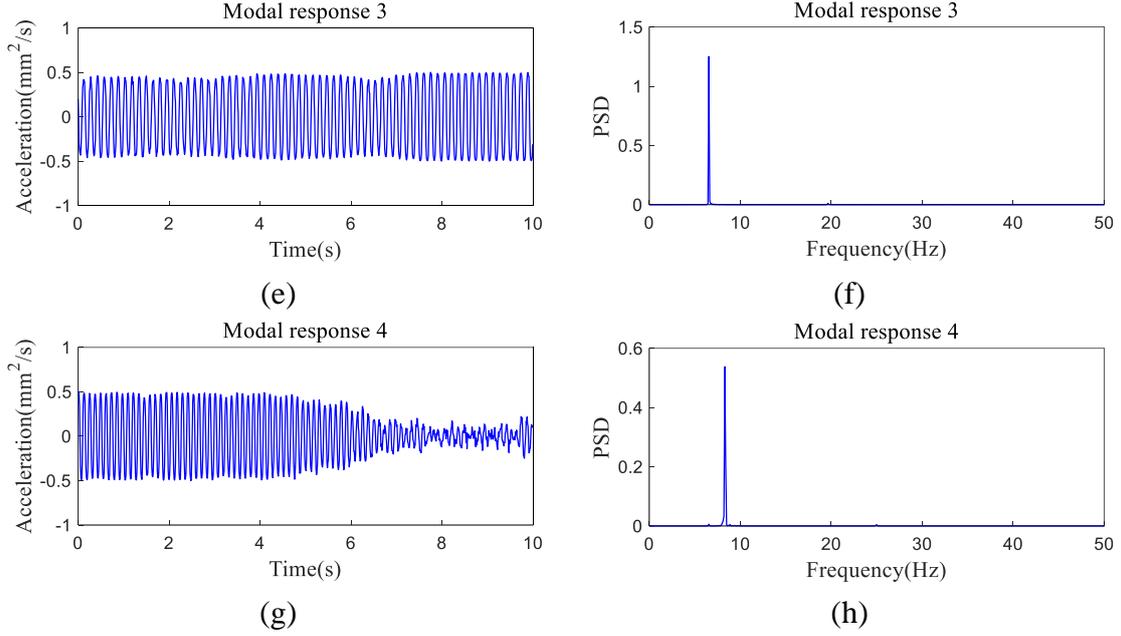

(e)                                              (f)

(g)                                              (h)

Figure 6 Estimated modal response results: (a), (b) first; (c), (d) second; (e), (f) third; and (g), (h) fourth modal responses

Modal parameters, including frequencies and damping ratios, are listed in Table 1. For comparison, the results identified by FDD, SSI, ICA, NExT+ERA, and ICA are also listed in Table 1. The results show that the frequencies and damping ratios are well identified by the proposed method, FDD, SSI, and NExT+ERA. The ICA method does not have good robustness, and the separation effect is very dependent on the data type; moreover, the ICA can only identify three modes.

Table 1 Identified modal parameters of the simulated system

| Mode | Frequency (Hz) | | | | | | Damping ratio (%) | | | | | |
|---|---|---|---|---|---|---|---|---|---|---|---|---|
| | Theoretical | FDD | SSI | ICA | NExT+ERA | Estimated | Theoretical | FDD | SSI | ICA | NExT+ERA | Estimated |
| 1 | 2.57 | 2.57 | 2.57 | - | 2.58 | 2.54 | 0.31 | 0.33 | 0.44 | - | 0.26 | 0.28 |
| 2 | 4.79 | 4.78 | 4.79 | 4.79 | 4.79 | 4.79 | 0.16 | 0.15 | 0.19 | 0.28 | 0.14 | 0.14 |
| 3 | 6.56 | 6.54 | 6.56 | 6.56 | 6.56 | 6.54 | 0.12 | 0.13 | 0.06 | 0.26 | 0.97 | 0.11 |
| 4 | 8.33 | 8.33 | 8.32 | 8.33 | 8.33 | 8.30 | 0.093 | 0.08 | 0.07 | 0.19 | 0.16 | 0.10 |

The results of the mode shapes are listed in Figure 7, the correlation between the estimated mode shape $\hat{\phi}_i$ $(i=1,\ldots,n)$ and theoretical mode shape $\phi_i$ is evaluated by the modal assurance criterion (MAC):

$$\mathrm{MAC}\left(\hat{\phi}_i \cdot \phi_i\right) = \frac{\left(\hat{\phi}_i^T \cdot \phi_i\right)^2}{\left(\hat{\phi}_i^T \cdot \hat{\phi}_i\right)\left(\phi_i^T \cdot \phi_i\right)} \tag{21}$$

The results in Table 2 show that the mode shapes are well identified by the proposed method, and all the MAC values are higher than 0.99. As an alternative modal identification method, the proposed machine-learning-based method has similar accuracy as the traditional FDD, SSI, and NExT+ERA methods.

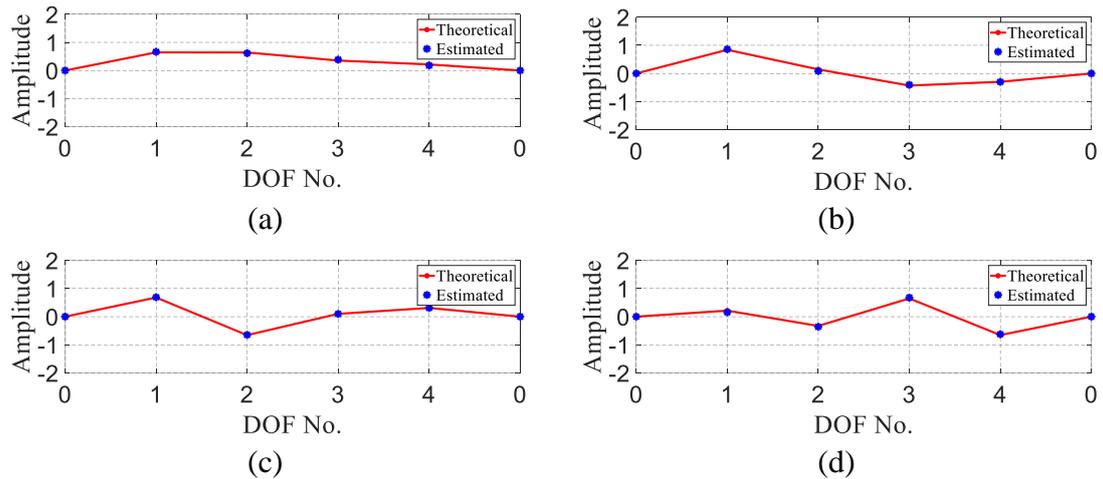

Figure 7 Identified and theoretical mode shapes: (a) first, (b) second, (c) third, and (d) fourth mode shapes

Table 2 MAC values of mode shapes

| Mode | MAC (FDD) | MAC (SSI) | MAC (NExT+ERA) | MAC (Estimated) |
| --- | --- | --- | --- | --- |
| 1 | 1.000 | 1.000 | 1.000 | 0.996 |
| 2 | 1.000 | 1.000 | 1.000 | 0.995 |
| 3 | 1.000 | 1.000 | 1.000 | 1.000 |
| 4 | 1.000 | 1.000 | 1.000 | 0.994 |

## 4. Example of a cable-stayed bridge

The vibration data from the structural health monitoring (SHM) system of an actual cable-stayed bridge were employed to verify the proposed method. As shown in Figure 8(a), the main bridge is a steel box-girder cable-stayed bridge with two towers and two cables. The total length of the bridge is 1,288 m, and the main span is 648 m. The SHM system was installed on this bridge in May 2006. There was a total of 18 accelerometers installed on the bridge main span and two side spans. In this example, 10 vertical accelerometers are selected as shown in Figure 8(b) to verify the proposed method's ability.

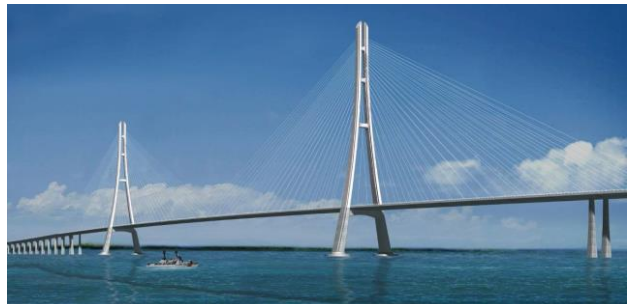

(a)

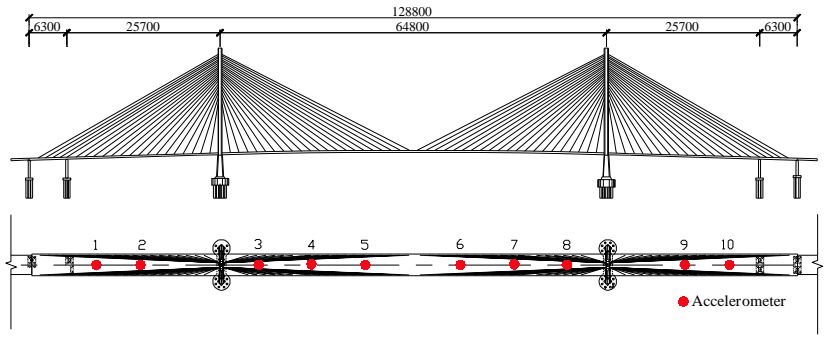

(b)

Figure 8 Actual bridge with SHM system: (a) bridge and (b) location of accelerometers

The acceleration data measured by the accelerometer sensors shown in Figure 8 are used to identify the modal parameters. The sampling frequency of the acceleration data is 10 Hz. In this example, data from 14:00 to 15:00 on September 26, 2011 from accelerometer sensors (a total of 10 sensors) were used. The time response acceleration data, PSD, and time-frequency analysis results obtained by short-time Fourier transform (STFT) are shown in Figure 9, which shows the bridge vibration energy almost focused in the range 0–1 Hz.

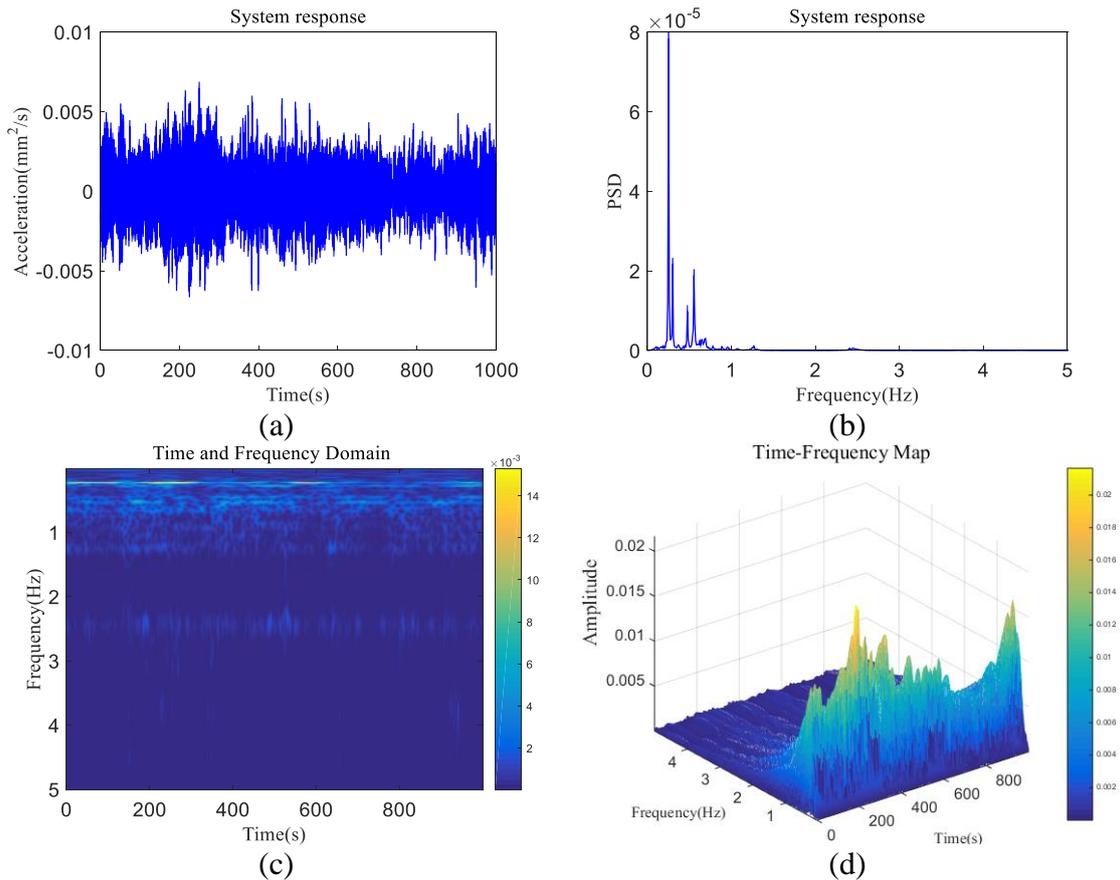

Figure 9 Original acceleration data :(a) time response data, (b) PSD, (c) time-frequency domain results, and (d) three-dimensional time-frequency distribution map

The architecture of the designed neural network is shown in Figure 10(a). The acceleration data with a length of 1,000 s, that is, 10,000 sample points from the selected 10 accelerometers were input into the network. Since the real data has lower energy modes that are difficult to separate, there will be false modal responses in the third layer, so the third layer is set more neurons than the number of modal responses (which can be equal to the number of input neurons). After the network training is completed, the accurate modal responses are selected from the third layer to form into a new third layer. Then, the new third and fourth layers are run using the loss function *L*4 to obtain the weights between the new third and fourth layers (i.e., the optimal estimation of mode shapes). The learning rate, batch size, and training epochs of the neural network are 0.001, 128, and 1,000, respectively. The training and convergence of the network are shown in Figure 10(b), which indicates that the convergence measured by the proposed loss function of Eq. (13) becomes optimal at epoch 100.

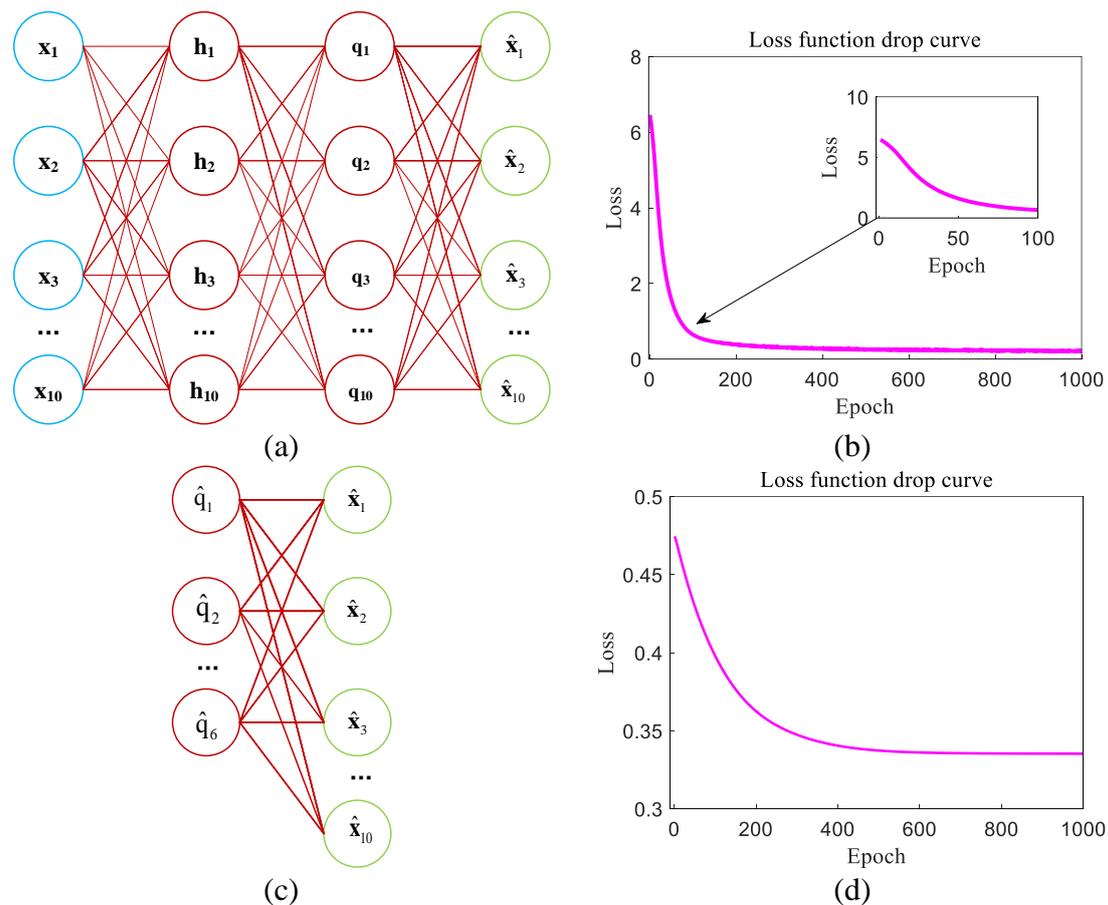

Figure 10 Neural network architecture and training results: (a), (c) network architecture and (b), (d) convergence in the training process

Ten modal responses separated by the neural network included the false modes. With the elimination of the false modes, the six modal responses are shown in Figure 11, in which the results in the time domain and the frequency domain show that the modal responses are accurately estimated. Using the selected six modal responses and running the new layers 3 and 4 of the neural network as shown in Figure 10(c), the coefficients of the mode shapes can be calculated. The corresponding loss function curve is shown in Figure 10(d).

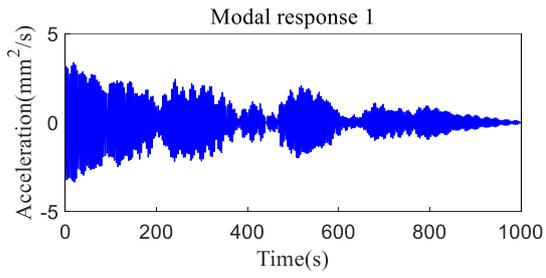 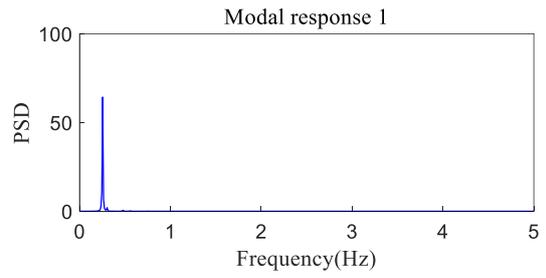

(a) (b)

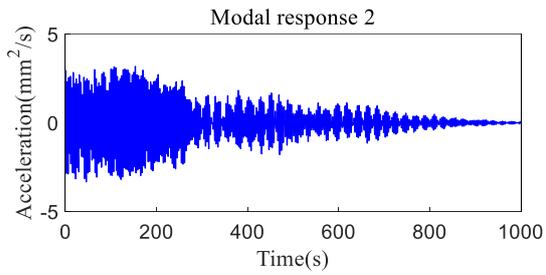 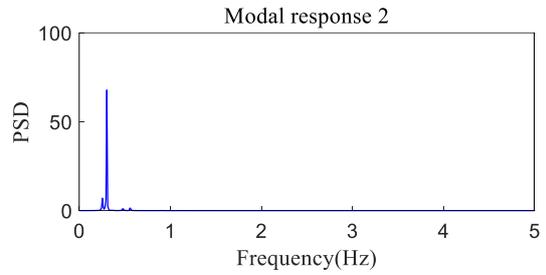

(c) (d)

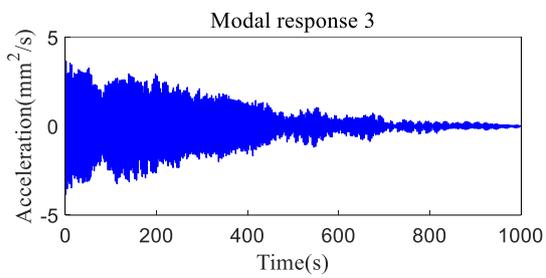 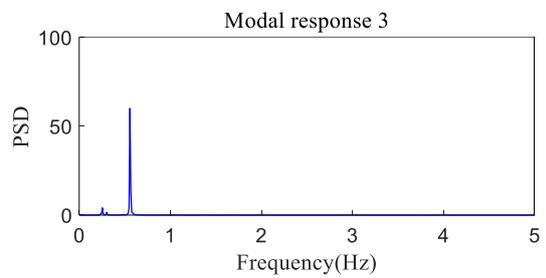

(e) (f)

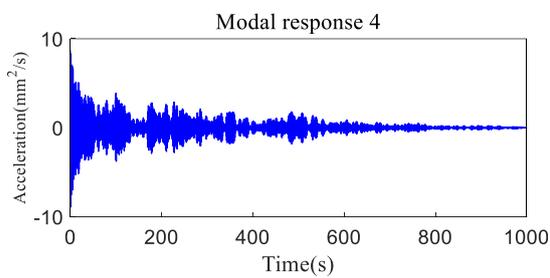 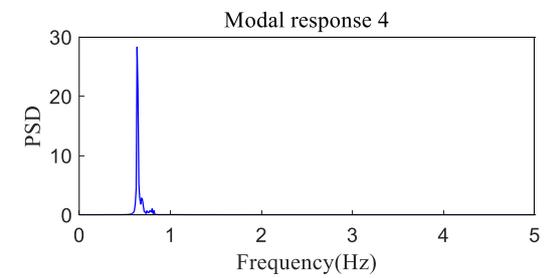

(g) (h)

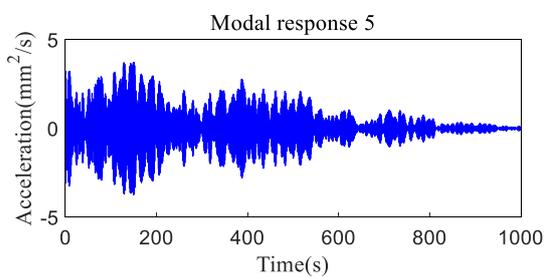 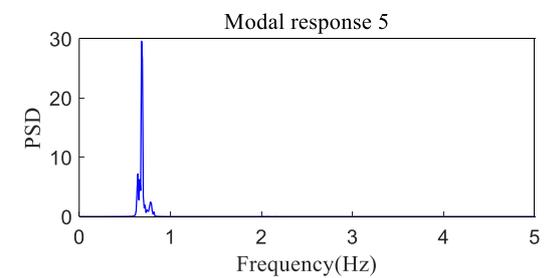

(i) (j)

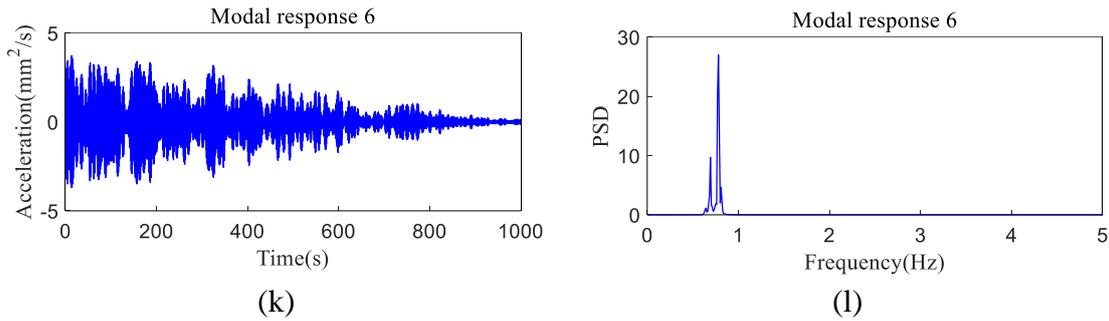

(k)                                       (l)

Figure 11 Modal response results: (a), (b) first; (c), (d) second; (e), (f) third; (g), (h) fourth; (i), (j) fifth; and (k), (l) sixth modal responses

Modal parameters, including frequencies and damping ratios, are listed in Table 3. For comparison, the frequency identified by finite-element, FDD, SSI, ICA, and NExT+ERA results, as well as the damping ratio identified by FDD, SSI, ICA, NExT+ERA results, is also listed.

Table 3 Identified modal parameters of the cable-stayed bridge

| Mode | Frequency (Hz) | | | | | | Damping ratio (%) | | | | |
|---|---|---|---|---|---|---|---|---|---|---|---|
| | Theoretical | FDD | SSI | ICA | NExT+ERA | Estimated | FDD | SSI | ICA | NExT+ERA | Estimated |
| 1 | 0.271 | 0.254 | 0.254 | 0.254 | 0.256 | 0.254 | 0.88 | 0.92 | 0.97 | 0.52 | 0.92 |
| 2 | 0.312 | 0.303 | 0.302 | - | 0.304 | 0.303 | 0.66 | 0.69 | - | 0.50 | 0.61 |
| 3 | 0.569 | 0.560 | 0.558 | 0.557 | 0.561 | 0.557 | 0.98 | 0.87 | 0.60 | 0.87 | 0.72 |
| 4 | 0.638 | 0.634 | 0.629 | - | 0.633 | 0.635 | 0.76 | 0.64 | - | 0.69 | 0.56 |
| 5 | 0.707 | 0.686 | 0.684 | - | 0.697 | 0.684 | 0.79 | 0.88 | - | 0.80 | 0.58 |
| 6 | 0.780 | 0.779 | 0.780 | 0.781 | 0.787 | 0.781 | 0.88 | 0.97 | 0.47 | 0.87 | 0.82 |

Except for the ICA method, the proposed neural network, FDD, NExT+ERA, and SSI methods all can identify the first six modes. The results further indicate that the proposed neural-network-based method can provide an alternative approach for structural modal identification under ambient excitation. The results of the mode shapes are listed in Figure 12, and the corresponding MAC values are shown in Table 4, which further confirms that the proposed method has a similar accuracy as the FDD, NExT+ERA, and SSI methods.

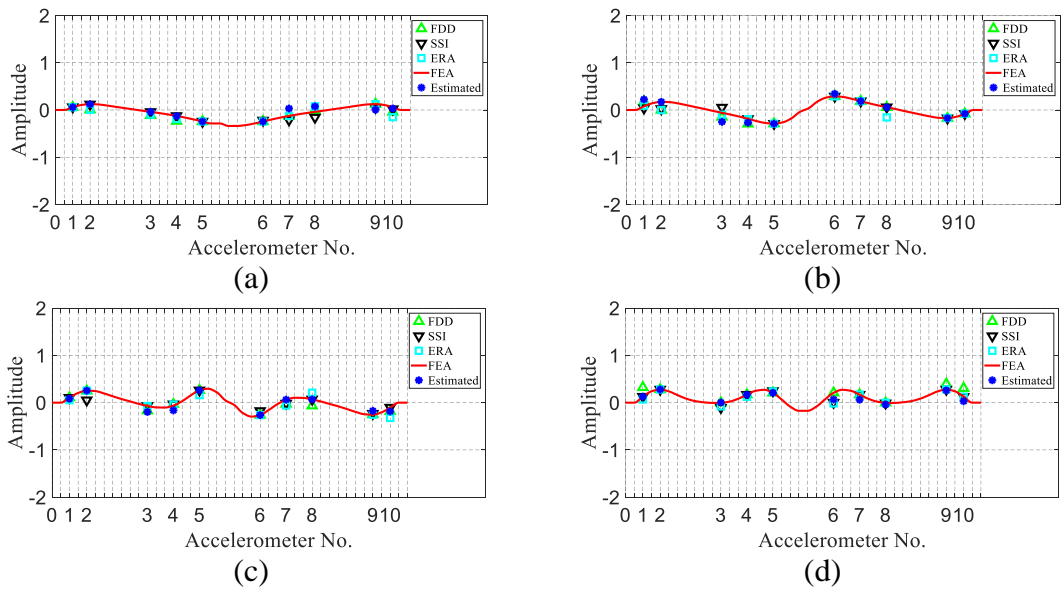

(a)                                       (b)

(c)                                       (d)

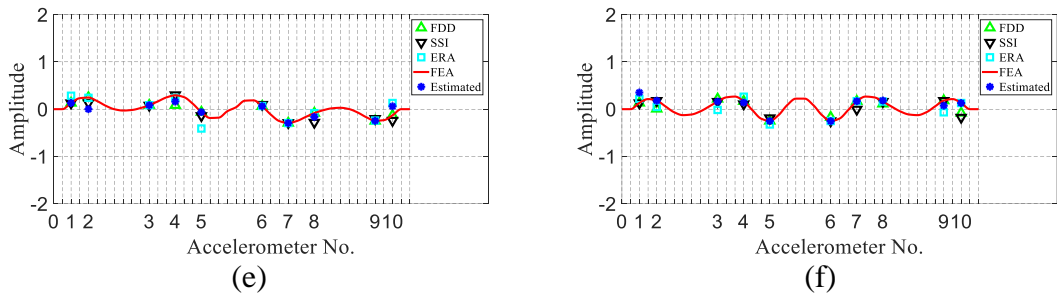

(e)  (f)

Figure 12 Estimated first six vertical mode shapes: (a) first; (b) second; (c) third; (d) fourth; (e) fifth; and (f) sixth

Table 4 MAC values of identified mode shapes

| Mode | MAC (FDD) | MAC (SSI) | MAC (NExT+ERA) | MAC (Estimated) |
|---|---|---|---|---|
| 1 | 0.8109 | 0.8663 | 0.6630 | 0.7305 |
| 2 | 0.8428 | 0.8853 | 0.7660 | 0.8799 |
| 3 | 0.8733 | 0.8338 | 0.7518 | 0.8841 |
| 4 | 0.8885 | 0.8418 | 0.8134 | 0.8772 |
| 5 | 0.8744 | 0.8272 | 0.6055 | 0.6549 |
| 6 | 0.7367 | 0.6198 | 0.6586 | 0.8534 |

## 5. Conclusions

In this study, a machine-learning-based approach for modal identification from the output-only data of structures was proposed. With the assumption of modal independence, the method first formalized the modal parameter identification problem into a DNN, and then constructed the loss function considering the uncorrelation and non-Gaussianity of modal responses to estimate the modal parameters. Verified by the numerical examples and the actual SHM data of a cable-stayed bridge, the results show that the proposed method has good modal identification accuracy similar to that of traditional FDD, SSI, and NExT+ERA methods, and obviously better than the ICA results, illustrating that the proposed method can provide an alternative approach for structural modal identification under ambient excitation.

In addition, the proposed method provides a new idea for modal identification in the framework of a DNN and fully exploits the powerful learning ability of neural networks in non-convex function optimization, which makes the modal identification more intelligent. Artificial intelligence (AI) will have wide application potential in SHM. The work described in this study should prove helpful in developing AI-based SHM data processing and structural safety assessment methods.

## Acknowledgment

This research was supported by grants from the National Key R&D Program of China (grant no. 2017YFC1500603), the National Natural Science Foundation of China (grant no. U1711265, 51678203 and 51638007, 51978216).